\documentclass[preprint]{article}
\usepackage{neurips_2022}
\usepackage{graphics}
\usepackage{graphicx}
\usepackage{booktabs}
\usepackage{amsmath}
\usepackage{pxfonts}
\usepackage{hyperref}
\newcommand{\comment}[1]{}

\title{Creative divergent synthesis with generative models}

\author{%
  Axel Chemla--Romeu-Santos, Philippe Esling \\
  IRCAM - Sorbonne Universit\'e - CNRS UMR 9912 STMS\\
  4, place Igor Stravinsky 75004 Paris\\
  \texttt{\{chemla, esling\}@ircam.fr}
}

\begin{document}

\maketitle

\begin{abstract}
Machine learning approaches now achieve impressive generation capabilities in numerous domains such as image, audio or video. However, most training \& evaluation frameworks revolve around the idea of strictly modelling the original data distribution rather than trying to extrapolate from it. This precludes the ability of such models to diverge from the original distribution and, hence, exhibit some creative traits. In this paper, we propose various perspectives on how this complicated goal could ever be achieved, and provide preliminary results on our novel training objective called \textit{Bounded Adversarial Divergence} (BAD).
\end{abstract}

\section{Introduction}
Generative machine learning models can now produce impressive artifacts, that are considered to be almost \textit{on par} with human creative productions \cite{esling2020creativity}.
However, besides introducing numerous problems in terms of ethics, copyrights, and aesthetics, these models are known to generally produce \textit{more-of-the-same} artifacts. This comes from the fact that models are mostly trained to reproduce existing works through \textit{reconstruction} tasks. Hence, such training setups strongly encourages the model to work in \textit{interpolation regimes}, requiring tremendous amounts of data to be able to accurately interpolate between different data modes. Conversely, we could think of more "creative" ways of using these architectures by trying to enforce them to \textit{extrapolate} outside of the input data distribution. In this paper, we introduce and experiment new directions to try to stimulate such behaviors by defining novel bounded divergence objectives between the data \& generative distributions through different types of formalization.

Approximating generative processes with machine learning tools is a non-trivial task, such that most research efforts addressed the ability of generative models to \textit{reproduce} a given target dataset. However, once state-of-art methods were deemed able to produce sufficiently valuable artifacts, some methods arose (mostly from the artistic community) to deviate models such as StyleGAN \cite{karras2020analyzing} from their original distribution. These methods were recently grouped under the potential name of \textit{active divergence} \cite{broad2021active}, encompassing different types of approaches. In a recent paper, \cite{chemla2022challenges} proposed several insights on how to stimulate such behaviors by divergence maximization. Yet, stimulating \textit{extrapolative} behaviors from generative models is a daunting task, and may not be achievable through a theoretical framework. Furthermore, evaluation of divergent generative models poses additional challenges, mostly due to the absence of reference data. Hence, such objectives not only requires the definition of novel training objectives, but also the development of meaningful quantitative evaluation metrics. 

\section{Approach and results} 
\label{sec:approach}
A naive approach for encouraging models to generate out-of-distribution samples would be to maximize the divergence between the target and generative distributions. However, such a direct "inverse" optimization would either most likely lead to a trivial and useless noise distribution. Similarly, starting from a pre-trained model would generally lead to \textit{catastrophic forgetting}, where the model would only produce low-quality samples without retaining any structural information about the original corpus. To circumvent such degenerate cases, a solution would rather be to design a novel min/max game of two different objectives. Given that we can compute how much a generation $\mathbf{x} \in \mathcal{X}$ belongs to a generic domain (e.g. images), and how much it belongs to a given set of classes (e.g. classes of digits), we could optimize our generic distribution to reinforce the first while reducing the second. This objective could then be written
\begin{equation}
\label{eq:bad}
    \mathrm{max}_\theta \ \mathbb{E}_{\mathbf{z} \sim p(\mathbf{z})} \big[ \alpha \mathcal{L}_{\mathrm{div}} \big( g_\theta(\mathbf{z}), \mathbf{x} \big) -  \beta \mathcal{L}_{\mathrm{reg}}\big(g_\theta(\mathbf{z}), \mathbf{x}\big) \big]
\end{equation}
where $g_\theta$ is a pre-trained parametric generator with latent prior $p(\mathbf{z})$, $\mathcal{L}_{\mathrm{div}}$ is a loss maximizing the distance to the chosen classes, $\mathcal{L}_{\mathrm{reg}}$ is a loss forcing the generative distribution to remain in the global distribution, and $(\alpha, \beta)$ are fixed weights. However, such criteria rely on high-level characteristics that may be tedious to retrieve directly from the data domain. Hence, we propose to compute these metrics using embeddings from pre-trained high-level classifiers, that we obtain from state-of-the-art or custom trainings. Regarding losses, we chose two different couples $\mathcal{L}_{\mathrm{div}}$ and $\mathcal{L}_{\mathrm{reg}}$, whose details are given in appendix sec.~\ref{sec:criteria}. This way, the optimization setup drives the model between two complementary objectives: maximizing the divergence with existing classes in the data, while being constrained to still belong to the overall target distribution. This setup, that we term \textit{Bounded Adversarial Divergence} (BAD), allows to enforce the model to generate coherent but novel samples.

\begin{figure}
    \centering
    \includegraphics[scale=0.31]{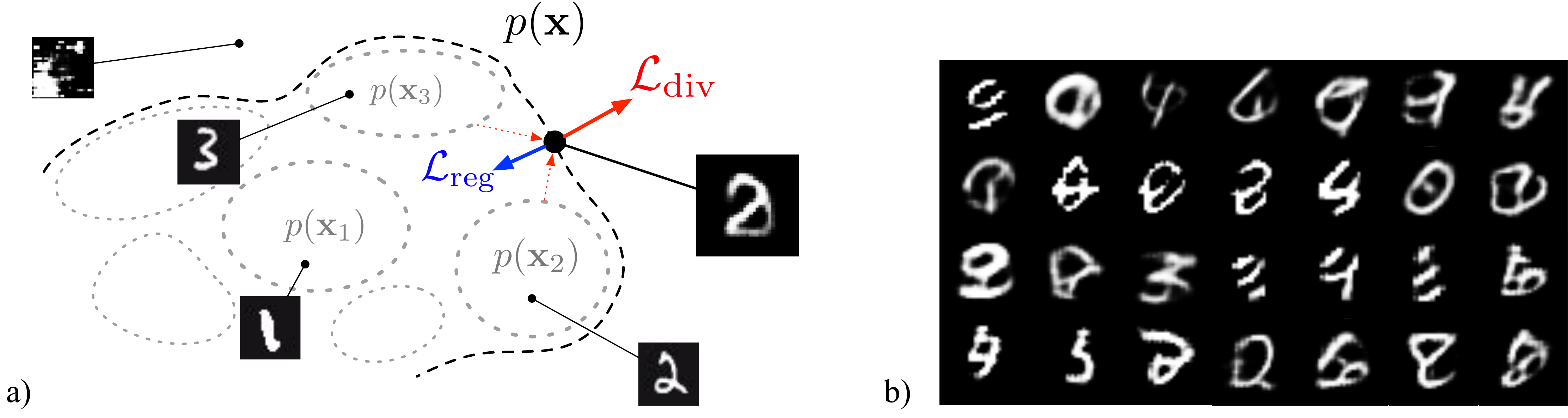}
    \caption{\small(a) Bounded Adversarial Divergence (BAD) is based on two competing losses: loss $\mathcal{L}_{\mathrm{div}}$ encourages the model to generate out-of-class samples, while loss $\mathcal{L}_{\mathrm{reg}}$ prevents it to generate outside of the overall target data distribution. (b) Cherry-picked generations from a fine-tuned VAE using the BAD framework ; full generations available in fig.~\ref{fig:generations}.}
    \label{fig:outcomes-main}
\end{figure}

To be able to intuitively evaluate the outcomes of this training procedure we focused on the MNIST dataset. Training procedures and results obtained with the different loss couples and classifiers are explained sec.~\ref{sec:training}. We also compute precision-recall \& diversity scores to provide qualitative measurements on how the model diverges from the original model and display the complete results in sec.~\ref{sec:evaluation}. 

\section{Conclusion}
In this paper, we proposed and implemented a training setup for generative models, called \textit{Bounded Adversarial Divergence} (BAD), that enforces a pre-trained model to diverge from its target distribution. We also provided preliminary results of this setup along quantitative measurements. This setup will be investigated more deeply in future works and applied to richer generative models in image, audio and video domains. 

\newpage
\bibliographystyle{abbrvnat}
\bibliography{main}

\newpage
\appendix

\section{Designing criteria for bounded adversarial divergence}
\label{sec:criteria}

As introduced in sec.~\ref{sec:approach}, the bounded adversarial divergence framework is grounded on the tension between two separate criteria : 
\begin{itemize}
    \item the divergent loss $\mathcal{L}_{\mathrm{div}}$, that pushes away the generative distribution from given data subsets,
    \item the regularization loss $\mathcal{L}_{\mathrm{reg}}$, that retains the generative distribution inside the overall data distribution.
\end{itemize}
However, obtaining such criteria directly from the data domain can be very tedious, mostly because of a high number of dimensions and examples, and their reliance on high-level data properties. A way to alleviate these issues is to rely on external class information $\mathbf{y}$, reflecting high-level attributes of the data, that could be used to derive our objective. To this end, we can retrieve internal embeddings from an external classifier as a feature space, provided it was trained on a similar data domain. Using such embeddings can then allow to both reduce the dimensionality for such measurements, and to guide the divergence with domain-specific high-level information. We propose two different setups $(\mathcal{L}_{\mathrm{div}}, \mathcal{L}_{\mathrm{reg}})$ that can be derived from such data embeddings.

\paragraph{Inception score and FID.}  The first $(\mathcal{L}_{\mathrm{div}}, \mathcal{L}_{\mathrm{reg}})$ is based on two different metrics : the \textit{Inception Score} (IS) and \textit{Frechet Inception Distance} (FID), two losses proposed to evaluate generative adversarial models \citep{salimans2016improved, heusel2017gans, barratt2018note}. IS is defined by
\begin{align*}
    \mathcal{L}_{\mathrm{IS}} = \exp{\mathbb{E}_\mathbf{x} D_{KL}\big[ p( \mathbf{y} | \mathbf{x} ) || p(\mathbf{y})\big]}
\end{align*}
that can be described as the divergence between the label distribution given a batch of generations $p(\mathbf{y|x})$, and the overall label distribution $p(\mathbf{y})$. As a high IS induces a high classification confidence for a given $\mathbf{x}$, minimizing it would enforce our divergent model to generate poorly classified outputs, considering each class as contrastive data subsets for our loss $\mathcal{L}_{\mathrm{div}}$. \\
FID is also a criterion coming from the generative model evaluation domain that rather intends to quantify the "quality" of the generation. To this end, it computes the Frechet distance between embedded features from real images and features from the model outcomes
\begin{align*}
    \mathcal{L}_{\mathrm{FID}} = |\mu - \mu_w| + tr\big(\Sigma + \Sigma_w - 2(\Sigma \Sigma_w)^{\frac{1}{2}}\big)
\end{align*}
where ($\mu$, $\Sigma)$ and $(\mu_w, \Sigma_w)$ are the two first moments of the data and generative distribution. Hence, a low $\mathcal{L}_{FID}$ means target and generative distributions globally match. Applying this criterion to our case, we can then measure how much the overall divergent distribution goes away from original data, and use FID as our regularization loss $\mathcal{L}_{\mathrm{reg}}$.

\paragraph{Maximum mean discrepancy.} Maximum Mean Discrepancy (MMD) is a derivable statistical two-sample test to measure the divergence between two distributions \cite{gretton2012kernel}. Given an appropriate positive-definite kernel $k: \mathcal{X} \times \mathcal{X} \rightarrow \mathbb{R}$, we can define the $MMD$
\begin{align*}
    MMD^2(\mathbf{x, y}) = \frac{1}{m(m-1)} \sum_{i, j\neq i} k(\mathbf{x}_i, \mathbf{x}_j) - 2 \frac{1}{m^2} \sum_{i, j} k(\mathbf{x}_i, \mathbf{y}_j) + \frac{1}{m(m-1)} \sum_{i, j\neq i} k(\mathbf{y}_i, \mathbf{y}_j)
\end{align*}
that can be defined as a distance in a Reproducing Kernel Hilbert Space. MMD is a widely used metric in machine learning, not only because of its symmetry but also because it is provides a computable way to compare two implicit distributions from samples. In this work, we relied on MMD to obtain both $\mathcal{L}_{\mathrm{div}}$ and $\mathcal{L}_{\mathrm{reg}}$. Taking the features obtained from the divergent model $f(\mathbf{x}_{\mathrm{div}})$ and the features obtained from original data $f(\mathbf{x}) = \{ f( \mathbf{x}_k )\}_{k\in[0, K]}$ (where $k$ is the corresponding digit class), we can obtain
\begin{align}
    \mathcal{L}_{\mathrm{div}} & = \sum_{k=1}^K MMD\big( f(\mathbf{x}_{\mathrm{div}}), f(\mathbf{x}_k) \big) \\
    \mathcal{L}_{\mathrm{reg}} & = MMD\big( f(\mathbf{x}_{\mathrm{div}}), f(\mathbf{x}) \big)
\end{align}
such that maximizing the classwise $\mathcal{L}_{\mathrm{div}}$ pushes the divergent distribution away from the class clusters, while minimizing the global MMD $\mathcal{L}_{\mathrm{reg}}$ enforces it to globally match the digit distribution.

\paragraph{Classifiers.} These divergences resort to internal representations of external classifiers as data embeddings. The chosen classifier is then of first importance, as it will drive how the generative distribution will actually diverge. In this work, we relied on two types of classifiers :
\begin{itemize}
    \item pre-trained generic classifiers available in the \texttt{torchvision} package \citep{marcel2010torchvision}: \texttt{alexnet}~\citep{krizhevsky2017imagenet}, \texttt{inception\_v3}~\citep{salimans2016improved} and \texttt{mobilenet\_v2}~\citep{howard2017mobilenets},
    \item a custom convolutional classifier,  described in Section~\ref{sec:training}.
\end{itemize}
The first classification models are trained on generic images with general classes, such that we can expect the resulted divergence to be less specific than the second, specifically trained on MNIST images with digit classes. By comparing these alternatives, we intend to see how much the chosen classifier impacted the resulting divergent distribution.

\paragraph{A note on divergence biases.} While we used these divergences to compute $\mathcal{L}_{\mathrm{div}}$ and $\mathcal{L}_{\mathrm{reg}}$, it is known that IS, FID and MMD are consequently biased when obtained from a low amount of samples \citep{barratt2018note}. The results obtained in the paper are obtained with a batch size that may be insufficient to obtained unbiased estimations of these divergences. We intend to resolve this limitation in future works.

\section{Results}
All the models used in this paper, besides the ones available in the \texttt{torchvision} package, were trained with the code at the following address :

\url{https://github.com/domkirke/divergent-synthesis}.\\

\subsection{Training details and architectures.}
\label{sec:training}

\paragraph{Obtaining the pre-trained VAE.} The generator used in
this work is the decoder of a \textit{variational auto-encoder} (VAE) \citep{kingma2013auto} with a 16-dimensional latent space, trained on the binary MNIST dataset \citep{lecun1998mnist}. The architecture of this VAE is described in Table~\ref{table:vae_dec}, with the encoder part having the reversed architecture. This model has been trained until convergence using a learning rate of $1\mathrm{e}^{-4}$ and batch size of 256 with the Adam optimizer \cite{kingma2014adam}, with a warm-up procedure from 0 to 1 in 10000 batches. Reconstruction errors, as well as the latent regularization term (Kullback-Leibler divergence between the encoder's outputs and the prior) are shown in Table~\ref{table:vae_results}. 

\paragraph{Obtaining the custom classifier.} The custom classifier is a simple 3-layered convolutional network, whose architecture is shown in Table~\ref{table:classifier}. Data augmentations were used to increase the classifier's generalization, without altering too much the nature of the digits : random affine transposition, random gaussian blur, and random erasing (with a single augmentation chosen with probability $0.5$). Regarding the optimization procedure, the classifier was trained on batches of size 128 using the Adam optimizer until convergence, with a learning rate of $1\mathrm{e}^{-3}$. Final classification results can be seen in Table~\ref{table:classifier_results}. 

\paragraph{Diverging from pre-trained generator.} Our divergent models are obtained by taking the decoder of the baseline VAE, decoding latent samples from the prior $\mathbf{z} \sim \mathcal{N}(\mathbf{0}, \mathbb{I})$ and optimizing jointly the proposed objective ~(\ref{eq:bad}) with random data batches. These models were optimized using the Adam optimizer with learning rate $5\mathrm{e}^{-6}$. The parameters $(\alpha, \beta)$ were selected to make the two adversarial losses equivalent, as they may depend on the chosen classifier. For the \texttt{torchvision} classifiers, $(\alpha, \beta)$ were set to $(-1.0, 0.1)$ for the IS/FID couple (IS has to be inverted, as we aim a low score) and $(1.0, 5.0)$ for the MMD. For the custom classifiers, weights were set to $(-1.0, 1.0)$ for the IS/FID and $(1.0, 20.0)$ for the MMD. Models were trained on 180 epochs on 40 randomly-selected batches, with a batch size of $256$ for the \texttt{torchvision} classifiers (due to memory constraints) and a size of $2048$ for the custom classifier. 

\begin{table}
\small
\parbox{.45\linewidth}{
  \caption{VAE decoder architecture}
  \label{table:vae_dec}
  \centering
  \begin{tabular}{|l|}
    Linear(16,800) \\ 
    BatchNorm1d() \\
    ELU() \\
    Linear(800,64) \\
    ConvTr2d(16, 32, ks=(9,9), st=(4,4), pad=(4,4)) \\
    ELU() \\
    ConvTr2d(32, 32, ks=(7,7), st=(2,2), pad=(3,3)) \\
    ELU() \\
    ConvTr2d(32, 64, ks=(5,5), st=(2,2), pad=(2,2)) \\
    ELU() \\
    ConvTr2d(64, 1, ks=(3,3), st=(1,1), pad=(1,1)) \\
    Sigmoid()
  \end{tabular}
  } 
  \hfill
  \parbox{.45\linewidth}{
  \caption{MNIST Classifier architecture}
  \label{table:classifier}
  \centering
  \begin{tabular}{|l|}
    Linear(16,800) \\ 
    BatchNorm1d() \\
    LeakyReLU() \\
    Linear(800,64) \\
    Conv2d(1, 64, ks=(3,3), st=(4,4), pad=(4,4)) \\
    LeakyReLU() \\
    Conv2d(64, 32, ks=(7,7), st=(2,2), pad=(3,3)) \\
    LeakyReLU() \\
    Conv2d(32, 16, ks=(9,9), st=(2,2), pad=(2,2)) \\
    Softmax() \\
  \end{tabular}
  } 
\end{table}

\begin{table}
\small
\parbox{.45\linewidth}{
  \caption{VAE results}
  \label{table:vae_results}
  \begin{tabular}{||l|l|l||}
    \textbf{Loss} & \textbf{Train} & \textbf{Validation} \\
    \midrule
    Cross-Entropy & 77.17 & 77.50\\
    Mean-squared Error & 23.98 & 24.06\\
    KLD & 23.61 & 23.66
  \end{tabular}
  } 
  \hfill
  \parbox{.45\linewidth}{
  \caption{Classification results}
  \label{table:classifier_results}
  \begin{tabular}{||l|l|l||}
    \textbf{Loss} & \textbf{Train} & \textbf{Validation} \\
    \midrule
    Cross-Entropy & 0.67 & 0.67\\
    Classification Ratio & 71.5\% & 71.3\%
  \end{tabular}
  } 
\end{table}

\subsection{Evaluations}
\label{sec:evaluation}
\begin{figure}
    \centering
    \includegraphics[width=0.9\textwidth]{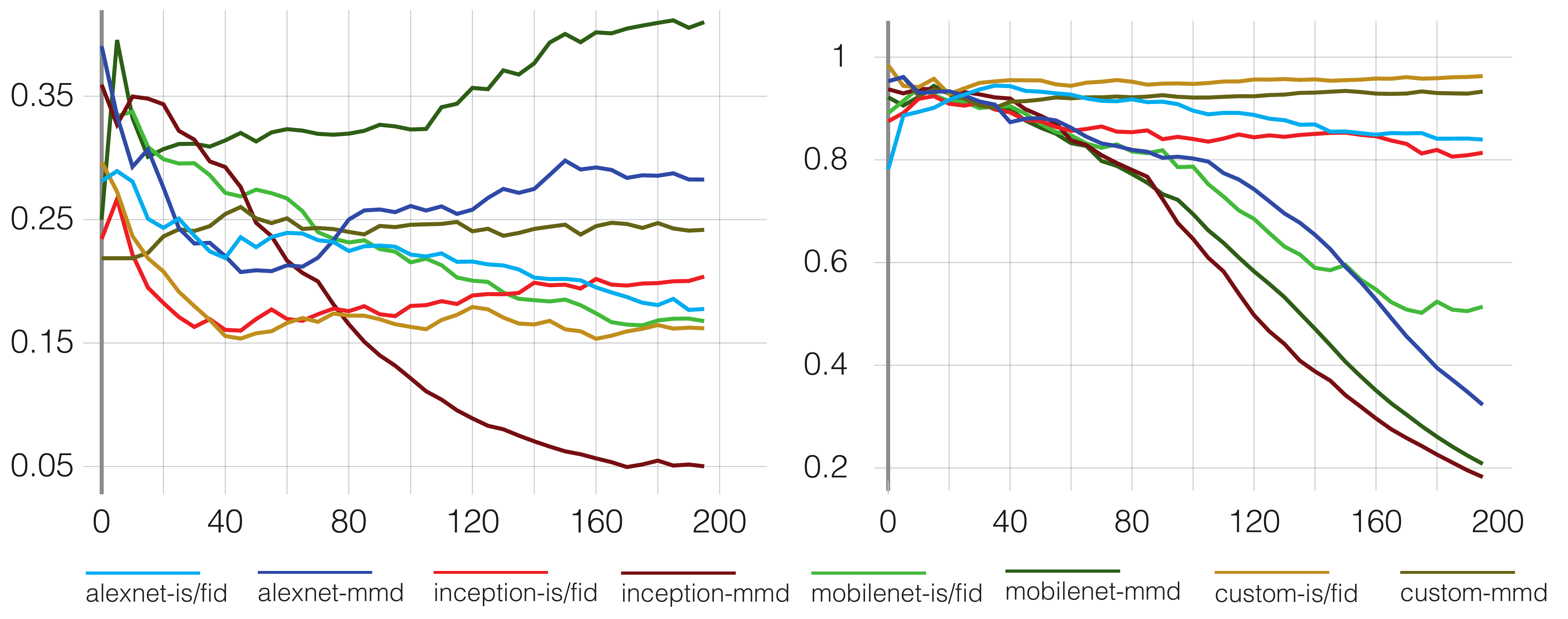}
    \caption{Precision-recall figures for different $(\mathcal{L}_{\mathrm{div}}, \mathcal{L}_{\mathrm{reg}})$ trade-offs.}
    \label{fig:evaluation-pr}
\end{figure}

A significant problem when addressing extensional cases in machine learning is how to evaluate the obtained generations. Indeed, without any reference data for outliers, how to quantify the quality of the obtained divergent generations can be quite paradoxical. Yet, the recent dawn of generative models fostered the development of evaluation procedures for generative models by the scientific community. Among these propositions, an interesting measure called \textit{precision-recall} proposes quantitative measurements for comparing a given generative distribution
$p_\theta(\mathbf{x})$ and its reference data $p(\mathbf{x})$ through the intersection of their relative supports \citep{sajjadi2018assessing}: while precision describes how much $p_\theta(\mathbf{x})$ is a subset of $p(\mathbf{x})$, recall describes how much $p(\mathbf{x})$ is included in $p_\theta(\mathbf{x})$. Formally, precision and recall are defined as scalars $(p, r)$ through distributions $\mu(\mathbf{x})$, $\nu_p(\mathbf{x})$ and 
$\nu_{p_\theta}(\mathbf{x})$ such that
\begin{align*}
    p(\mathbf{x}) & = r \mu(\mathbf{x})  + (1 - r) \nu_p(\mathbf{x})  \\
    p_\theta(\mathbf{x})  & = p \mu(\mathbf{x})  + (1 - p) \nu_{p_\theta}(\mathbf{x}) 
\end{align*}
With this definition, distributions $\nu_p(\mathbf{x}), \nu_{p_\theta}(\mathbf{x})$ are respectively the part of $p(\mathbf{x})$ missed by $p_\theta(\mathbf{x})$, and the original information brought by $p_\theta(\mathbf{x})$. These measures can then provide us qualitative measurements on how much our divergent distributions overlaps with the original VAE. The evolution of these criteria for several prior temperatures (standard deviation of the centered normal used for latent sampling) are available in Figure~\ref{fig:evaluation-pr}.

\begin{figure}
    \label{fig:generations}
    \centering
    \includegraphics[width=0.9\textwidth]{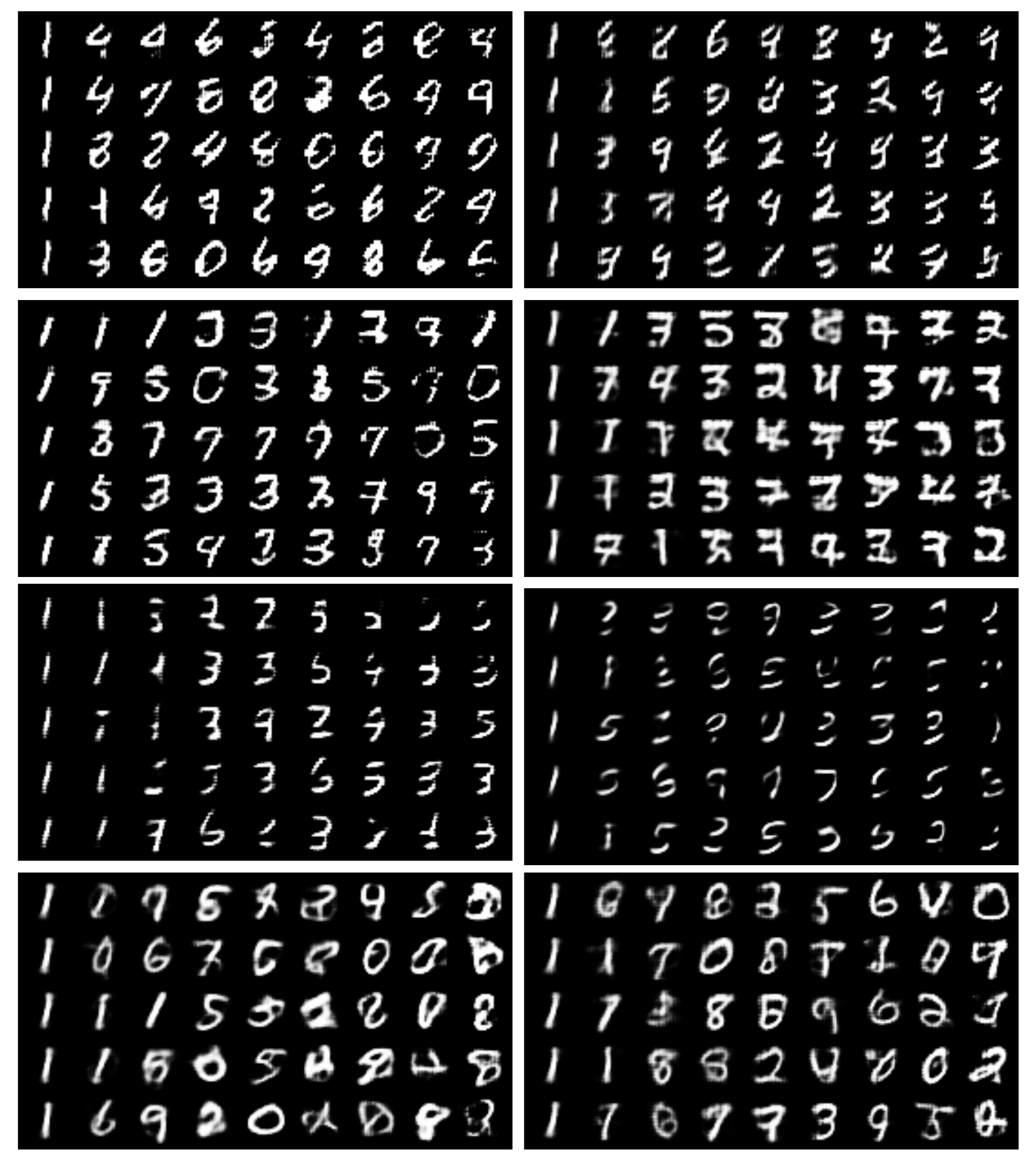}
    \caption{Generations obtained with IS/FID (left) and MMD (right) divergences. From top to bottom, generations obtained with : \texttt{alexnet}, \texttt{inception\_v3}, \texttt{mobilenet\_v2}, and our custom classifier. Within each image, columns are generations obtained with different temperatures~ : $[0.1, 0.5, 0.7, 1.0, 1.2, 1.5, 1.8, 2.0, 3.0]$. Epochs were cherry-picked, but full generations are available on our github.}
\end{figure}

\section {Ethics}
All the results shown above were produced using a minimal setup, involving a single GPU and light generative/classification models. We think that fostering more research on "extrapolative" machine learning is promising for several reasons. First, investigating how much generative models can deviate from their original distribution provides alternative insights on the real features learned by the model, but also stray these models from their imitating behavior that pose cultural issues in terms of originality, artistic interest, and copyrighting. Second, we think that extrapolative setups may also be fruitful in low data regimes, providing methods for modeling rich generative distributions from a few examples, hence, lowering the energy cost of these models for creative applications.

\end{document}